\documentclass{article}

\usepackage[final]{corl_2019} 
\usepackage{url}
\usepackage{multicol}
\usepackage{graphicx}
\usepackage[font=footnotesize]{caption}
\usepackage{subcaption}

\usepackage{amsthm}
\usepackage{amsmath, amssymb}
\usepackage{bbm}
\usepackage{bm}
\usepackage{siunitx}
\usepackage{dsfont}
\usepackage{wrapfig}

\newcommand{\norm}[1]{\left\lVert #1 \right\rVert}

\title{Reinforcement Learning without Ground-Truth State}

\author{
  Xingyu Lin\hspace{5mm}Harjatin Singh Baweja\hspace{5mm}David Held\\
  Robotics Institute\\
  Carnegie Mellon University\\
  \texttt{\{xlin3, harjatis, dheld\}@andrew.cmu.edu} \\
}

\begin{document}
\maketitle


\begin{abstract}

%
To perform robot manipulation tasks, a low-dimensional state of the environment typically needs to be estimated. However, designing a state estimator can sometimes be difficult, especially in environments with deformable objects. An alternative is to learn an end-to-end policy that maps directly from high-dimensional sensor inputs to actions. However, if this policy is trained with reinforcement learning, then without a state estimator, it is hard to specify a reward function based on high-dimensional observations. To meet this challenge, we propose a simple indicator reward function for goal-conditioned reinforcement learning: we only give a positive reward when the robot's observation exactly matches a target goal observation. We show that by relabeling the original goal with the achieved goal to obtain positive rewards~\cite{andrychowicz2017hindsight}, we can learn with the indicator reward function even in continuous state spaces. We propose two methods to further speed up convergence with indicator rewards: reward balancing and reward filtering.  We show comparable performance between our method and an oracle which uses the ground-truth state for computing rewards.  We show that our method can perform complex tasks in continuous state spaces such as rope manipulation from RGB-D images, without knowledge of the ground-truth state.
\end{abstract}

\keywords{Self-supervised, goal-conditioned reinforcement learning} 


\section{Introduction}
\setlength{\intextsep}{0pt}%
\begin{wrapfigure}{r}{0.4\textwidth}
  \begin{center}
    \includegraphics[width=\linewidth]{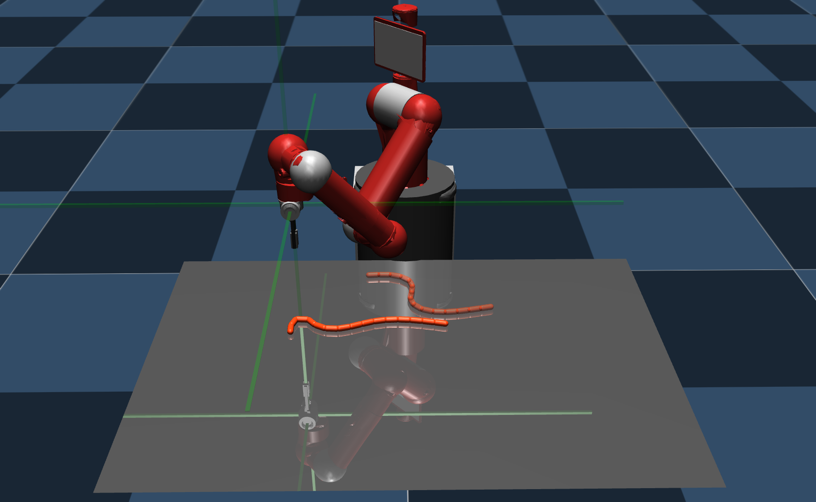}
  \end{center}
  \caption{An illustration of the rope pushing task. The Sawyer robot is given an image of the current configuration of the rope and an image of the goal configuration (illustrated as the  translucent rope) and the task is to push the rope to the goal configuration.}
  \label{fig:intro}
\end{wrapfigure}

To perform robot manipulation tasks, a low dimensional state of the environment typically needs to be estimated. In reinforcement learning, this state is also used to compute the reward function. However, designing a state estimator can be difficult, especially in environments with deformable objects, as shown in Figure \ref{fig:intro}. An alternative is to learn an end-to-end policy that maps directly from high-dimensional sensor input to actions. However, without a state estimator, it is hard to specify a reward function based on high-dimensional observations.

Past efforts to use reinforcement learning for robotics have avoided this issue in a number of ways.  One common approach is to use extra sensors to determine the state of the environment during training, even if such sensors are not available at test time. Examples of this include using another robot arm to hold all relevant objects~\cite{levine2016end}, placing an IMU sensor~\cite{gu2017deep,yahya2017collective} or motion capture markers on such objects~\cite{kormushev2010robot}, or ensuring that all relevant objects are placed on scales~\cite{schenck2016guided}.

However, such instrumentation is not always easy to set up for each task.  This is especially true for deformable object manipulation, such as rope or cloth manipulation, in which every part of the object must be instrumented in order to measure the full state of the entire object.  Attaching such sensors to food or granular material would present additional difficulties. 


We present an alternative approach for goal-conditioned reinforcement learning for specifying rewards using raw (e.g. high-dimensional and continuous) observations without requiring explicit state estimation or access to the ground-truth state of the environment.  We achieve this using a simple indicator reward function, which only gives a positive reward when the robot's observation exactly matches a target goal observation.  Naturally, in continuous state spaces, we do not expect any two observed states to be identical. Surprisingly, we show that we can learn with such an indicator reward, even in continuous state spaces, if we use goal relabeling~\cite{Kaelbling93b,andrychowicz2017hindsight}, which relabels the original goal with the achieved observation such that a positive reward is given. As the indicator rewards produce extreme sparse positive rewards, we further introduce reward balancing to balance the positive and negative rewards, as well as reward filtering to filter out uncertain rewards.

We show theoretically that the indicator reward results in a policy with bounded suboptimality compared to the ground-truth reward.  We also empirically show comparable performance between our method and an oracle which uses the ground-truth state for computing rewards, even though our method only operates on raw observations and does not have access to the ground-truth state.  
We demonstrate that an indicator reward can be used to teach a robot complex tasks such as rope manipulation from RGB-D images, without knowledge of the ground-truth state during training. Videos of our method can be found at \url{https://sites.google.com/view/image-rl}.

\section{Related Work}
\subsection{Obtaining Ground-truth State for Training}
\textbf{Adding sensors: }
To obtain ground-truth states for calculating rewards, one approach is to perform state estimation.  However, such an approach can be noisy and challenging to implement, especially for the deformable objects that we study in this work.  Another approach is to add extra sensors during training to accurately record the state. For example, in past work, one robot arm (covered with a cloth at training time) is used to rigidly hold and move an object, while another robot arm learns to manipulate the object~\cite{levine2016end}.  In such a case, the object position can be inferred directly from the position of the robot gripper that is holding it.  In other work on teaching a robot to open a door, an IMU sensor is placed on the door handle to determine the rotation angle of the handle and whether or not the door has been opened~\cite{gu2017deep,yahya2017collective}.   One can also ensure that all relevant objects for a task are placed on scales~\cite{schenck2016guided} or affixed with motion capture markers to obtain a precise estimate of their position~\cite{kormushev2010robot}.  However, such instrumentation is challenging for deformable objects, granular material, food, or other settings. Further, such instrumentation is costly and time-consuming to setup; hence most of these previous approaches assume that such instrumentation is only available at training time and these methods do not allow further fine-tuning of the policy after deployment.


\textbf{Training in simulation: } Another common approach is to train the policy entirely in simulation in which ground-truth state can be obtained from the simulator.~\cite{fang2018learning,andrychowicz2018learning,zhu2018reinforcement,sadeghi2016cad2rl,pinto2018asym}.  Many approaches have been explored to try to transfer such a policy from simulation to the real world, such as domain randomization \cite{tobin2017domain} or building a more accurate simulator~\cite{tan2018sim,chebotar2018closing}.  However, obtaining an accurate simulator is often very challenging, especially if the simulator differs from the real-world in unknown ways. Further, building the simulator itself can be fairly complex.  Because these methods require the ground-truth state to obtain the reward function, they require training in a simulator and do not allow further fine-tuning after deployment in the real world; our method, in contrast, does not require the ground-truth state for the reward function.




\subsection{Robot Learning without Ground-truth State}

\textbf{Learning a reward function without supervision: }
One line of work for learning a reward function is to first learn a latent representation and then derive the reward function based on the distance in the embedding space, such as cosine similarity. The representation can be learned by maximizing the mutual information between the achieved goal and the intended goal~\cite{Wardefarley2018}, reconstruction of the observation with VAE~\cite{nair2018visual}, or learning to match keypoints with spatial autoencoders~\cite{finn2016deep}. Recent work also explicitly learns a representation that is suitable for gradient based optimizer and then use it for specifying rewards \cite{yu2019unsupervised}. However, there is no guarantee that these learned representation are suitable for deriving rewards for control. In addition, if the representation is pre-trained, it may also be incorrect in some parts of the observation space which can be exploited by the agent. Our approach is much simpler in that the reward function does not have any parameters that need to be learned and we empirically show better performance to some reward learning approaches.


\textbf{Changing the optimization:}
Another approach is to forego maximizing a sum of rewards as is typically done in reinforcement learning and instead optimize for another objective.  For example, one method is to choose one-step greedy actions based on a learned one-step inverse dynamics model; after training, the policy is then applied directly to a multi-step goal~\cite{agrawal2016learning}.
An alternative method is to learn a predictive forward dynamics model directly in a high-dimensional state space and use visual model-predictive control~\cite{finn2016unsupervised,finn2017deep,ebert2017self,ebert2018robustness,ebert2018visual}.  Although these methods have shown some promise, predicting future high-dimensional observations (such as images or depth measurements) is challenging.  Another approach is to obtain expert demonstrations and define an objective  as trying to imitate the expert~\cite{sermanet2018time,sermanet2016unsupervised,peng2018variational,finn2017one}. Our approach, however, applies even when demonstrations are not available.


\subsection{Manipulating Deformable Objects} 
Deformable object manipulation presents many challenges for both perception and control.  One approach to the perception problem is to perform non-rigid registration to a deformable model of the object being manipulated~\cite{huang2015leveraging,lee2015non,schulman2013tracking,wang2011perception,javdani2011modeling,miller2011parametrized,cusumano2011bringing,phillips2014representation}.  However, such an approach is often slow, leading to slow policy learning, and can produce errors, leading to poor policy performance.  Further, such an approach often requires a 3D deformable model of the object being manipulated, which may be difficult to obtain.  Our approach applies directly to high-dimensional observations of the deformable object and does not require a prior model of the object being manipulated.




\section{Problem Formulation}
In reinforcement learning, an agent interacts with the environment over discrete time steps. In each time step $t$, the agent observes the current state $s_t$ and takes an action $a_t$. In the next time step, the agent transitions to a new state $s_{t+1}$ based on the transition dynamics $p(s_{t+1} | s_t, a_t)$ and receives a reward $r_{t+1} = r(s_t, a_t, s_{t+1})$. The objective for the agent is to learn a policy $\pi(a_t|s_t)$ that maximizes the expected future return $R = \mathbb{E}\big[ \sum_{t=0}^\infty \gamma^t r_{t+1} \big]$, where $\gamma$ is a discount factor.

\subsection{Goal-reaching Reinforcement Learning} In order for the agent to learn diverse and general skills, we define a goal reaching MDP~\cite{schaul2015universal,andrychowicz2017hindsight} as follows: In the beginning of each episode, a goal state $s_g$ is sampled from a goal distribution $\mathcal{G}$. We learn a goal conditioned policy $\pi(a_t|s_t, s_g)$ that tries to reach any goal state from the goal distribution. We use a goal conditioned reward function $r_t = r(s_{t+1}, s_g)$ and optimize for $\mathbb{E}_{s_g\sim \mathcal{G}}\big[ \sum_{t=0}^\infty \gamma^t r_t \big]$. The transition dynamics $p(s_{t+1}| s_t, a_t)$ of the environment remain independent of the goal. 

In many real-world scenarios, it is often difficult to construct a well-shaped reward function. Past work has shown that sparse rewards, combined with an appropriate learning algorithm, can achieve better performance than poorly-shaped dense rewards in goal-reaching environments~\cite{andrychowicz2017hindsight}.
We thus define a sparse reward function that only makes the binary decision of whether the goal is reached or not. Specifically, let $S_+(s_g)$ be a subset of the state space such that any state in this set is determined to be sufficiently close to $s_g$ (in some unknown metric); in other words, if the environmental state is within $S_+(s_g)$, then the task of reaching $s_g$ can be considered to be achieved.\footnote{$S_+$ is a function that maps from the state space to a subset of the space.} Naturally, we can assume that $s_g \in S_+(s_g)$.
A binary reward function can then be defined as 
\begin{equation}
 r(s_{t+1}, s_g)= 
    \begin{cases} 
        R_+ &  s_{t+1} \in S_{+}(s_g)\\
        R_- &  s_{t+1} \notin S_{+}(s_g),
    \end{cases}
 \label{eqn:true_reward}
\end{equation}
where $R_+$ and $R_-$ are constants representing the rewards received for achieving the goal and failing to achieve the goal, respectively. 

\subsection{Rewards from Images}
In many cases, the ground-truth state $s_t$ is unknown and we cannot directly use the true reward function defined in Equation~\ref{eqn:true_reward}.  Instead, the agent observes  high-dimensional observations $o_t$ from sensors, from which we must instead define a proxy reward function $\hat{r}(o_{t+1}, o_g)$. The question now becomes how to choose $\hat{r}$ to be optimal for reinforcement learning? 


The most common approach in robotics is to perform state estimation.  However, in many cases, the state estimator might be hard to obtain, such as for deformable object manipulation, e.g. laundry folding or food preparation.
We therefore investigate whether an alternative reward function that does not depend on state estimation can be used. Specifically, let us consider a general reward function defined in observation space of the form
\begin{equation}
 \hat{r}(o_{t+1}, o_g)= 
    \begin{cases} 
        R_+ &  o_{t+1} \in \hat{O}_{+}(o_g)\\
        R_- &  o_{t+1} \notin \hat{O}_{+}(o_g),
    \end{cases}
 \label{eqn:noisy_reward}
\end{equation}
where $o_g$ is a representation of the goal in observation space  and $\hat{O}_{+}(o_g)$ is a subset of the observation space for which we will give positive rewards.  A number of past approaches have proposed various methods for learning an observation-based reward function $\hat{r}(o_{t+1}, o_g)$~\cite{Wardefarley2018, nair2018visual, florensa2019self, yu2019unsupervised}. However, these approaches do not analyze the properties needed by such a reward function to enable optimal learning.  Next we will investigate trade-offs between different choices of $\hat{O}_{+}(o_g)$ and how they will affect policy training time when trained with rewards of $\hat{r}(o_{t+1}, o_g)$.

\section{Reward Misclassifications}

\setlength{\intextsep}{0pt}%
\begin{wrapfigure}{r}{0.65\textwidth}
  \begin{center}
    \includegraphics[width=\linewidth]{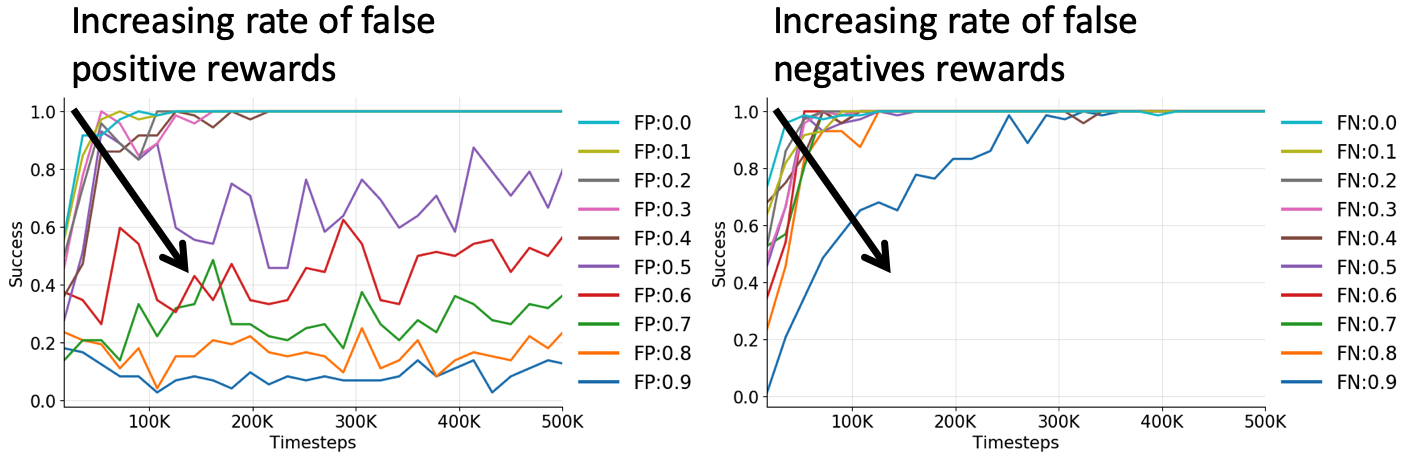}
  \end{center}
  \vspace*{-3mm}
  \caption{As we increase false negative/positive rewards, the learning curves with false positive rewards are affected more severely.} 
\label{fig:reward_flipping}
\end{wrapfigure}

We will now investigate how to design a good proxy reward function $\hat{r}(o_{t+1}, o_g)$, based on raw sensor observations, that we can use to train the policy; we desire for the policy trained with $\hat{r}(o_{t+1}, o_g)$ to optimize the original reward $r(s_{t+1}, s_g)$ based on the ground-truth state (which we do not have access to).  Our first insight into choosing a good proxy reward function $\hat{r}(o_{t+1}, o_g)$ is that we should think about reward functions in terms of false positives and false negatives.  Let us define a false positive reward to occur when the agent receives a positive reward based on our proxy reward function $\hat{r}(o_{t+1}, o_g)$ when it would have received a negative reward based on the original reward function $r(s_{t+1}, s_g)$.  In other words, suppose that an unknown function $f$ maps from an observation to its corresponding ground-truth state, $s_t = f(o_t)$.  Then a false positive reward occurs when  $o_{t+1} \in \hat{O}_+(o_g)$ while $f(o_{t+1}) \notin S_+(s_g)$. Similarly, a false negative reward can be defined.



Intuitively, both false positive rewards and false negative rewards can negatively impact learning.  However, for any estimated reward function $\hat{r}(o_{t+1}, o_g)$, we will have either false positives or false negatives (or both) unless we have access to a perfect state estimator. To design a good proxy reward function, we must ask: which will more negatively affect learning: false positives or false negatives?



The two types of mistakes are not symmetric. As we will see, a false positive reward can significantly hurt policy learning, while a false negative reward is much more tolerable. Under a false positive reward, the agent receives a positive reward (under the proxy reward function $\hat{r}(o, o_g)$) for reaching some observation $o$, even though the agent should receive a negative reward based on the corresponding ground-truth state $f(o)$ under the original reward function $r(s_t, s_g)$.  
This false positive reward will encourage the agent to continue to try to reach the state $f(o)$, even though reaching this state does not achieve the original task since $f(o) \notin S_+(s_g)$.  





On the other hand, false negative rewards are much more tolerable.  Under a false negative, the agent observes some observation $o$ such that $f(o) \in S_+(s_g)$ but the agent receives a negative reward.  However, if the agent still receives a positive reward for some other observation $o'$ such that $f(o') \in S_+(s_g)$, then the agent can still learn to reach the goal states $S_+(s_g)$, though learning might be slower and the learned policy may be suboptimal.
 


We provide a simple example to verify this intuition. Consider a robot arm reaching task, with the observation space  $\mathcal{O} \in \mathcal{R}^3$ being the 3D position of the end-effector (EE). The action space $\mathcal{A} \in \mathcal{R}^3$ controls the position of EE. The true reward is defined by $O_+(o_g) = \{o\mid \norm{o-o_g}_2 < \epsilon \}$. We define two types of noisy reward functions used for training. The reward function $\hat{r}_{FP}$ gives the same rewards as the true reward function, except that with a probability of $p_{FP}$ (False Positive Rate), a negative reward will be flipped to a positive reward. The reward function $\hat{r}_{FN}$ can be similarly defined, where a positive reward will be flipped to a negative reward with a probability of $p_{FN}$ (False Negative Rate). For this experiment, we use a standard reinforcement learning algorithm DDPG \cite{lillicrap2015continuous} combined with goal relabeling \cite{andrychowicz2017hindsight}. The learning performance of this same algorithm with different noisy rewards can be shown in Figure \ref{fig:reward_flipping}. We can see that the agent is able to learn the task even with a very large false negative rate. But when the false positive rate increases, the performance sharply decreases.



\section{Approach} \label{sec:approach}
\subsection{Indicator Rewards}
Following this idea, we propose using a proxy reward function that does not have any false positive rewards.   To do so, we will use an extreme reward function of $\hat{O}_+(o_g) = \{o_g\}$.  In other words, we will use an indicator reward function:
\begin{equation}
 \hat{r}_{ind}(o_{t+1}, o_g)= 
    \begin{cases} 
        R_+ &  o_{t+1} =o_g\\
        R_- &  o_{t+1} \neq o_g,
    \end{cases}
 \label{eqn:indicator_reward}
\end{equation}
It should be clear that this reward function will have no false positives, since the reward is positive only if $o_{t+1} = o_g$, which implies that $f(o_{t+1}) = f(o_g)$, or equivalently, $s_{t+1} = s_g$.  As $s_g \in S_+(s_g)$ by definition, all positive rewards are true positives. However, this reward function is extremely sparse and has many false negatives.  In fact, without goal relabeling, in continuous state spaces, we would expect all rewards to be negative under this indicator reward function, as no two observations in continuous spaces will ever be identical.  Next, we will describe how to learn with this reward function with goal relabeling.



\subsection{Goal Relabeling for Off-policy Learning} 
%

Fortunately, for off-policy multi-goal learning, we can adopt the goal relabeling technique introduced in \cite{Kaelbling93b,andrychowicz2017hindsight} to learn the goal-conditioned Q-function. Suppose that some transitions $(o_t, a_t, o_{t+1})$ are observed when the agent takes
an action $a_t \sim \pi(o_t, o_g)$ with a goal of $o_g$.  Because Q-learning is an off-policy reinforcement learning algorithm, we can replace the goal observation $o_g$ with any other observation $o_{g'}$ in our Bellman update of the Q-function. This works because the transition dynamics $p(s_{t+1}|s_t, a_t)$, and likewise the observation transition dynamics $p(o_{t+1}| o_t, a_t)$, are independent of the goal $o_g$.  Specifically, for some transitions, we will choose to replace $o_g$ with the observation
$o_{t+1}$.  By re-labeling $o_g$ with $o_{t+1}$ and using our indicator reward function, we will have that $\hat{r}_{ind}(o_{t+1}, o_g) = \hat{r}_{ind}(o_{t+1}, o_{t+1}) = R_+$. Thus, using goal relabeling, we can get positive rewards, even when using an indicator reward function in continuous state spaces.



\subsection{Reward Balancing and Filtering}
As mentioned above, after sampling a batch of data, we train the Q-function with goal relabeling.  We use three different strategies for choosing which goals to use for relabeling: with probability $p_1$, we relabel $o_g$ with $o_{t+1}$, which will receive a positive reward under our indicator reward function.  With probability $p_2$, we relabel the goal $o_g$  with $o_{t'}$ with an observation from some future time $t'$ step within the episode. The indicator reward function will most likely give a negative reward in this case, which is possibly a false negative because the new goal is possibly considered achieved based on the state-based, ground-truth reward funciton. Finally, with probability $p_3$, we use the original goal (with no relabeling), which will again most likely give a negative reward under the indicator reward function; as before, this might be a false negative.

\textbf{Reward balancing: } We refer to ``reward balancing" as setting $p_1 = p_2 = 0.45$ and $p_3 = 0.1$, leading us to receive positive rewards approximately $0.45$ of the time and negative rewards approximately $0.55$ of the time.  Thus the ratio of positive and negative rewards that we use to train the Q-function are approximately balanced, even with indicator rewards. From another perspective, $p_1$ and $p_2$ determine the relative frequency between providing positive rewards and propagating rewards to other timesteps in the episode. Additionally, training with a small fraction of the original goals (i.e. $p_3$) can be seen as a regularization which ensures that the distribution of the relabeled goals moves towards the original goal distribution.

\textbf{Reward filtering:} While false negative rewards do not hurt learning as much as false positives, we still wish to avoid them if possible to improve the convergence time of the learned policy. We achieve this using ``reward filtering," in which we filter out transitions that we suspect of having a high chance of being false negatives. We refer to ``reward filtering" as discarding a sampled transition if its Q value is above a threshold $q_0$. If the assigned reward is negative based on the proxy reward function but the Q-value is sufficiently high, then there is a chance that this reward a false negative.  To reduce the fraction of false negatives, we filter out such transitions and do not use them for training. 

We can estimate how to set the threshold $q_0$ as follows: for a given transition $(o_t, a_t, o_{t+1})$, if $o_{t+1} = o_g$, we know that $\hat{r}_{ind}(o_{t+1}, o_g) = R_+$. In this case, $Q^*(o_t, a_t, o_g) = R_+/(1-\gamma)$, where $Q^*$ is the optimal Q-function, assuming the optimal policy will continue to receive (discounted) positive rewards in the future. Similarly, if $o_{t+1} \neq o_g$, then $\hat{r}_{ind}(o_{t+1}, o_g) = R_-$. Since we know that the policy starting from $o_t$ will thus receive at least one negative reward before receiving positive rewards, then $Q^*(o_t, a_t, o_g) \leq R_- + \gamma R_+/(1-\gamma)$. Thus, we can set a threshold $q_0$, where $R_- + \gamma R_+/(1-\gamma) < q_0 < R_+/(1-\gamma)$; if we find that $Q(o_t, o_g, a_t) > q_0$, then the corresponding reward $\hat{r}_{ind}(o_{t+1}, o_g)$ is likely to be a false negative (assuming that the Q-function has been trained well); we thus filter out such rewards, to reduce the number of false negatives that we use for training. We can see that $q_0$ is set to a rather conservative fitlering value. Additionally, the Q-function is initialized to a relatively low value to avoid overestimation of the Q-function which can lead to incorrect filtering in the beginning of the training.

\section{Analysis}
In this section, we analyze the performance of learning with indicator rewards. We first interpret the goal conditioned Q-function as a measure of the time it takes for the agent to reach one observation from another.
\subsection{Minimum Reaching Time Interpretation} \label{sec:min_reaching}
Let us define $d = D_\pi(o_t, a_t, \hat{O}_+(o_g))$ as the number of time steps it takes for the policy $\pi$ to go from the current observation $o_t$, starting with action $a_t$, to reach the set $\hat{O}_+(o_g)$ of goal observations. For simplicity, we assume that, once the agent receives a positive reward, it will take actions to continue to receive positive rewards. The Q-function can be written as 
\begin{equation}
    \begin{split}
    Q_\pi(o_t, a_t, o_g) &= R_-+ \gamma R_- + ... + \gamma^{d-1}R_- + \gamma ^d R_+ + \gamma ^{d+1} R_+ + ...\\
        &= \frac{\gamma^d}{1-\gamma} (R_+ - R_-) + \frac{1}{1-\gamma} R_-.
    \end{split}
\end{equation}
Now it can be easily seen that, as long as $R_+ > R_-$, $Q_{\pi}$ is strictly monotonically decreasing w.r.t. $d$. As such, maximizing $Q_\pi$ over $\pi$ is equivalent to minimizing the time the agent takes to reach the goal $\hat{O}_+(o_g)$. Note that this is true for varying definitions of $\hat{O}_+(o_g)$; thus the policy trained under the true reward function (Equation~\ref{eqn:true_reward}) will minimize $D_\pi(o_t, a_t, O_+(o_g))$ whereas the policy trained under the indicator reward function will minimize $D_\pi(o_t, a_t, \{o_g\})$ (slightly overloading notation for $D_\pi$).  Below we will show how this interpretation of the policy's behavior at convergence can lead to a simple analysis of the suboptimality of the learned policy under the indicator reward.





\subsection{Analysis of Sub-optimality}
Due to the false negative rewards given by the indicator function $\hat{r}_{ind}$, the learned policy may not be optimal with respect to the original reward function $r(s_{t+1}, s_g)$ defined in Equation~\ref{eqn:true_reward}. Here we give the worst case bound for the policy learned with the indicator reward.
Following the minimum reaching time interpretation of the previous section, we evaluate the performance of the policy in terms of the time it takes to reach the set of goal observations $O_+(o_g)$ from the current observation. Given $o_t, o_g$, denote $t_1$ as the minimum number of time steps to reach from $o_t$ to the set of true goal observations, i.e. $t_1 = D(o_t, O_+(o_g)) $. Let $t_2 = D(o_t, o_g)$ be the minimum time to reach from $o_t$ to $o_g$. Define the diameter of this goal observation set as $d = max\{D(o_1, o_2) | o_1, o_2 \in O_+(o_g) \}$. From the optimality of $t_2$, we know that
$$t_2 \leq D(o_t, o) + D(o, o_g), \forall o\in O_+(o_g).$$
Thus, 
\begin{equation}
    \begin{split}
        t_2 & \leq \min_{o\in O_+(o_g)} D(o_t, o) + D(o, o_g) \leq t_1 + d.
    \end{split}
\end{equation}
From the analysis in the previous section, the optimal policy which optimizes the indicator reward will reaches $o_g$ in $t_2$ time steps; since $o_g \in O_+(o_g)$, we know that this policy will reach $O_+(o_g)$ in some time $t_3 \leq t_2$.  Also recall that we have defined $t_1$ such that the optimal policy under the true reward function of Equation~\ref{eqn:true_reward} will reach $O_+(o_g)$ in $t_1$ steps. Thus $t_3/t_1 \leq (t_1 + d)/t_1$ is an upper bound on the suboptimality of the policy trained under the indicator reward, at convergence.

\section{Experiments}
Our experiments address the following questions:
\begin{enumerate}
    \item In the case of visual input, how much are the sample efficiency and the final performance affected without assuming access to the ground-truth reward?
    \item How much does reward balancing and filtering improve learning efficiency?
    \item Can our method scale to real world robotic tasks?
\end{enumerate}

We denote our method, which uses indicator rewards with reward balancing and filtering, as \textbf{Indicator+Balance+Filter}. We compare our method with the following baselines:
\begin{itemize}
    \item \textbf{Oracle}: This method 
    assumes access to the ground-truth reward from state space $r(s_t, s_g)$.
    \item \textbf{Indicator}: This is an ablation of our method, without reward balancing and filtering.
    \item \textbf{Auto Encoder (AE)}: We train an autoencoder with an $L2$ reconstruction loss of the image observation, jointly with the RL agent. We then use cosine similarity in the learned embedding space to provide dense rewards, as similarly compared in \cite{Wardefarley2018}. Specifically, assuming the learned encoding of an observation $o$ is $\phi(o)$ after $L2$ normalization, the reward will be $r(o, o_g) = max(0, \phi(o)^T\phi(o_g))$.
    \item \textbf{Variational Auto Encoder (VAE)}: Similarly to the AE baseline, a VAE is jointly trained with the RL agent to provide rewards, as done in \cite{nair2018visual}. For a fair comparison, the goal sampling strategy for this baseline is kept the same a other approaches.
    \item \textbf{Distributional Planning Network (DPN) \cite{yu2019unsupervised}}: DPN aims to learn a representaiton that is suitable for a gradient based planner to reach a goal observation. Following \cite{yu2019unsupervised}, we first pre-train the DPN using samples collected from a random policy and then use the learned representation for giving rewards. Note that in the plots below, we do not count these samples used for the pre-training.
\end{itemize}
Only Oracle uses the ground-truth, state-based reward function. We use the standard off-policy learning algorithm DDPG \cite{lillicrap2015continuous} with goal relabeling \cite{andrychowicz2017hindsight}. For methods without reward balancing, we re-label the current goal with an achieved goal sampled uniformly from one of the future time steps within the same episode with a probability of 0.9; otherwise, the original goals are used. For all the environments, the ground-truth rewards are based on the $L_2$ distance in the state space:
$R_+$ if  $\norm{s_{t+1} - s_g} \leq \epsilon$ and $R_-$ otherwise. More details on algorithms, architectures, and hyperparameters can be found in the Appendix.

We first evaluate all the methods in a set of simulated environments in MuJoCo \cite{todorov2012mujoco}, where both current and goal observation are given by RGB-D images:
\begin{itemize}
    \item Reacher: Teach a two-link arm to reach a randomly located position in 2D space.
    \item FetchReach: Move the end effector of the Fetch robot to a random position in 3D.
    \item RopePush: Push a rope from a random initial configuration into a target configuration.
\end{itemize}

The first two environments above are standard environments from Gym \cite{OpenAI2018}. For the more complex RopePush task, the robot needs to push a 15-link rope to a targeted pose, as shown in Figure \ref{fig:intro}. To accelearate learning, we fix the orientation of the gripper and parameterize the action as $(x_1, y_1, x_2, y_2) \in \mathcal{R}^4$, denoting the starting and ending position of one push from the gripper. We generate the initial rope pose by giving the rope a random push from a fixed location. The goal poses are generated by giving the rope two more pushes based on the initial push (these pushes are hidden from the policy). The robot can give three pushes to push the rope to the goal pose. More details on environments can be found in Appendix A.

The results are shown in Figure \ref{fig:visual_all_experiments}. We see that our method (Indicator+balance+Filter) achieves nearly the same performance as the Oracle even though our method operates only from RGB-D images and does not have access to the ground-truth state.  For the RopePush environment, only the Oracle and our method are able to learn to achieve the task to a reasonable accuracy. For AE, VAE and DPN, the learned representation may not lead to a perfect reward function everywhere and the agent will exploit states that yield a high proxy rewards, even though the goal is not achieved. Instead, our method does not require learning a reward function and outperforms these baselines.
\vspace{1mm}

\begin{figure*}[h]
    \centering
    \begin{subfigure}[t]{0.8\linewidth}
        \centering
        \includegraphics[width=\linewidth]{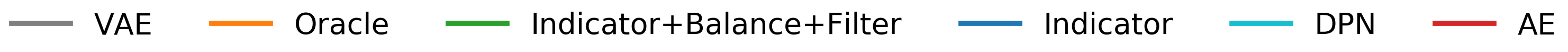}
    \end{subfigure}
    \begin{subfigure}[t]{0.32\linewidth}
        \includegraphics[width=\linewidth]{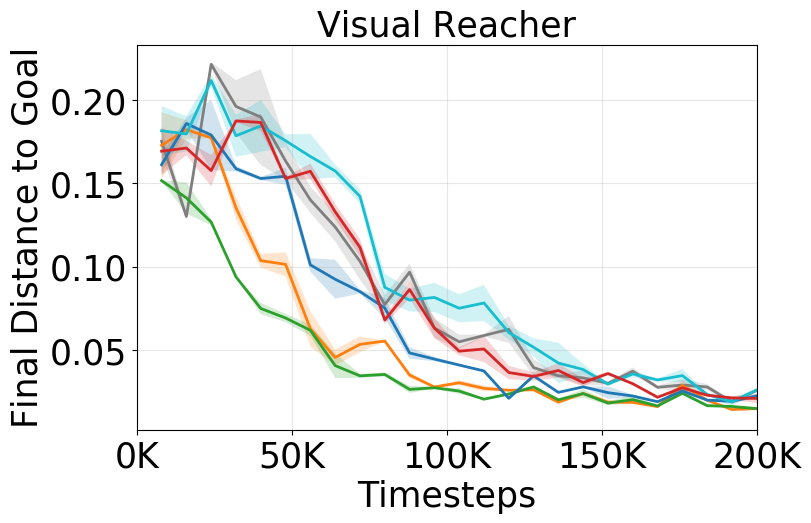}
        \label{fig:visual_Reacher_dist}
    \end{subfigure}
    \begin{subfigure}[t]{0.32\linewidth}
        \includegraphics[width=\linewidth]{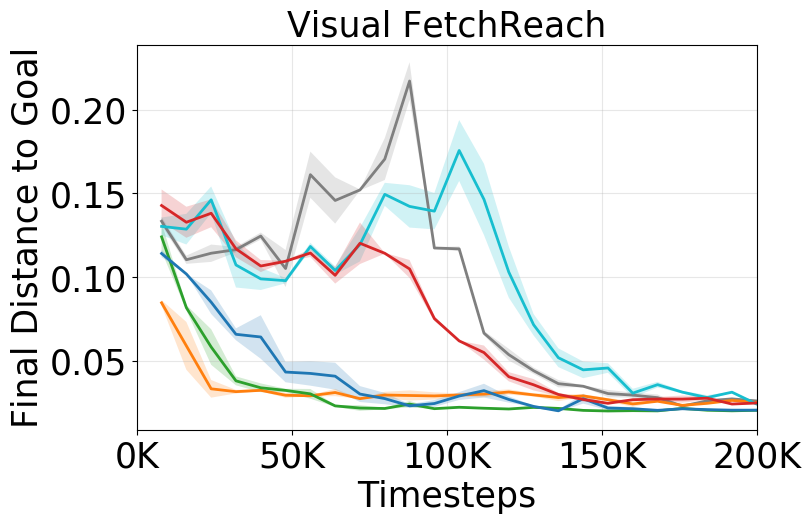}
        \label{fig:visual_FetchReach_dist}
    \end{subfigure}
    \begin{subfigure}[t]{0.3\linewidth}
        \includegraphics[width=\linewidth]{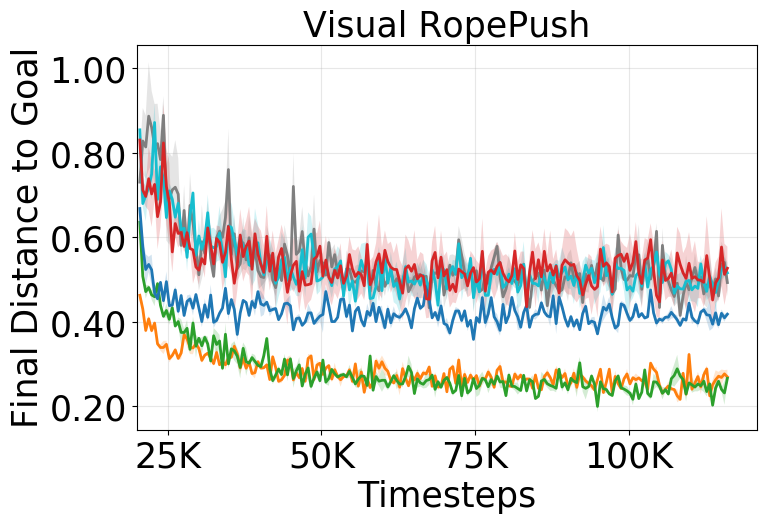}
        \label{fig:visual_RopeFloat_dist}
    \end{subfigure}
    \vspace*{-3mm}
    \caption{The final distance to goal of different methods in different environments throughout the training. The observations are from the RGB-D images rendered in simulation.}
    \label{fig:visual_all_experiments}
\end{figure*}

In Appendix C, we perform an ablation where we removed either Balance or Filter from our method.  These experiments show that both Balance and Filter are important for optimal performance across the environments tested.  
In Appendix D, we show experiments in which the ground-truth states are used as inputs to the policy but are not used to compute the rewards.  In these experiments, we also see that our method has a similar performance to the Oracle and outperforms the other approaches.


\begin{wrapfigure}{r}{0.5\textwidth}
    \begin{minipage}{0.5\textwidth}
    \centering
    \begin{subfigure}[t]{0.36\linewidth}
        \includegraphics[width=\textwidth]{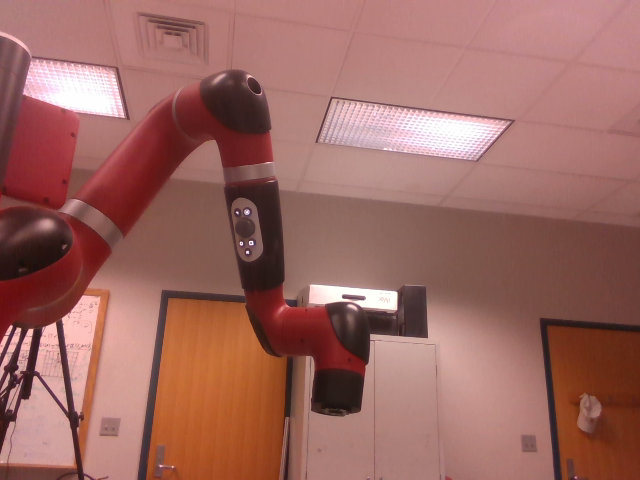}
        \label{fig:visual_sawyer_Reacher_obs}
    \end{subfigure}
    \begin{subfigure}[t]{0.36\linewidth}
        \includegraphics[width=\textwidth]{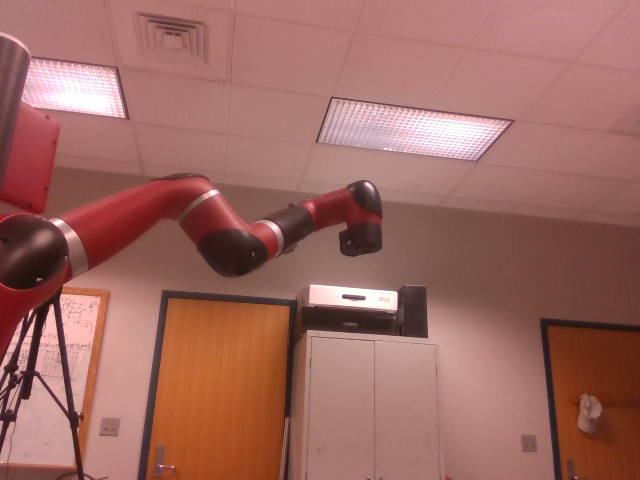}
    \end{subfigure}
    \begin{subfigure}[t]{0.56\linewidth}
        \includegraphics[width=\textwidth]{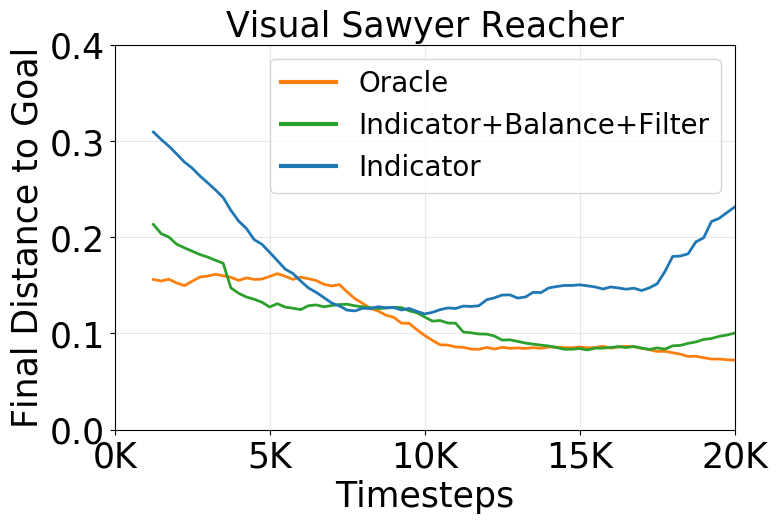}
        \label{fig:visual_sawyer_Reacher_dist}
    \end{subfigure}
    \vspace*{-6mm}
    \caption{An example observation image (top left) and goal image (top right); final distance to goal (bottom). }
    \label{fig:sawyer_visual_setup}
    \end{minipage}

\end{wrapfigure}



\subsection{Indicator Rewards with Real Images}
Using RGB-D observations and goals, we train a Sawyer robot for a 3-dimensional reaching task. Figure~\ref{fig:sawyer_visual_setup} shows example observation and goal images as well as the distance to goal throughout training. As before, our method (Indicator+Balance+Filter) performs similarly to the Oracle in terms of  final goal distance; the baseline of Indicator Rewards without balancing or filtering performs significantly worse and diverges in the end due to many negative rewards, many of which are false negative.

\section{Conclusion}
In this work, we show that we can train a robot to perform complex manipulation tasks directly from high-dimensional images, without requiring access to the ground-truth state in either the policy input or the reward function.  We empirically show that our method enables a robot to learn complex skills for manipulating deformable objects, for which state estimation is often challenging. We also demonstrate  that our method performs well in the real world.

We provide a theoretical analysis which shows that the optimal policy under the indicator reward has a bounded suboptimality compared to the optimal policy under the ground-truth reward. We hope that our method will enable robot learning in the real world in cases where it is difficult to add extra sensors or accurately simulate the environment.


\clearpage


\bibliography{ind_reward}  

\begin{thebibliography}{42}
\providecommand{\natexlab}[1]{#1}
\providecommand{\url}[1]{\texttt{#1}}
\expandafter\ifx\csname urlstyle\endcsname\relax
  \providecommand{\doi}[1]{doi: #1}\else
  \providecommand{\doi}{doi: \begingroup \urlstyle{rm}\Url}\fi

\bibitem[Andrychowicz et~al.(2017)Andrychowicz, Wolski, Ray, Schneider, Fong,
  Welinder, McGrew, Tobin, Abbeel, and Zaremba]{andrychowicz2017hindsight}
M.~Andrychowicz, F.~Wolski, A.~Ray, J.~Schneider, R.~Fong, P.~Welinder,
  B.~McGrew, J.~Tobin, O.~P. Abbeel, and W.~Zaremba.
\newblock Hindsight experience replay.
\newblock In \emph{NIPS}, pages 5048--5058, 2017.

\bibitem[Levine et~al.(2016)Levine, Finn, Darrell, and Abbeel]{levine2016end}
S.~Levine, C.~Finn, T.~Darrell, and P.~Abbeel.
\newblock End-to-end training of deep visuomotor policies.
\newblock \emph{The Journal of Machine Learning Research}, 17\penalty0
  (1):\penalty0 1334--1373, 2016.

\bibitem[Gu et~al.(2017)Gu, Holly, Lillicrap, and Levine]{gu2017deep}
S.~Gu, E.~Holly, T.~Lillicrap, and S.~Levine.
\newblock Deep reinforcement learning for robotic manipulation with
  asynchronous off-policy updates.
\newblock In \emph{ICRA}, pages 3389--3396. IEEE, 2017.

\bibitem[Yahya et~al.(2017)Yahya, Li, Kalakrishnan, Chebotar, and
  Levine]{yahya2017collective}
A.~Yahya, A.~Li, M.~Kalakrishnan, Y.~Chebotar, and S.~Levine.
\newblock Collective robot reinforcement learning with distributed asynchronous
  guided policy search.
\newblock In \emph{IROS}, pages 79--86. IEEE, 2017.

\bibitem[Kormushev et~al.(2010)Kormushev, Calinon, and
  Caldwell]{kormushev2010robot}
P.~Kormushev, S.~Calinon, and D.~G. Caldwell.
\newblock Robot motor skill coordination with em-based reinforcement learning.
\newblock In \emph{IROS}, pages 3232--3237. IEEE, 2010.

\bibitem[Schenck and Fox(2016)]{schenck2016guided}
C.~Schenck and D.~Fox.
\newblock Guided policy search with delayed sensor measurements.
\newblock \emph{arXiv preprint arXiv:1609.03076}, 2016.

\bibitem[Kaelbling(1993)]{Kaelbling93b}
L.~P. Kaelbling.
\newblock Learning to achieve goals.
\newblock In \emph{Proceedings of the Thirteenth International Joint Conference
  on Artificial Intelligence}, Chambery, France, 1993. Morgan Kaufmann.

\bibitem[Fang et~al.(2018)Fang, Zhu, Garg, Kurenkov, Mehta, Fei-Fei, and
  Savarese]{fang2018learning}
K.~Fang, Y.~Zhu, A.~Garg, A.~Kurenkov, V.~Mehta, L.~Fei-Fei, and S.~Savarese.
\newblock Learning task-oriented grasping for tool manipulation from simulated
  self-supervision.
\newblock \emph{RSS}, 2018.

\bibitem[Andrychowicz et~al.(2018)Andrychowicz, Baker, Chociej, Jozefowicz,
  McGrew, Pachocki, Petron, Plappert, Powell, Ray,
  et~al.]{andrychowicz2018learning}
M.~Andrychowicz, B.~Baker, M.~Chociej, R.~Jozefowicz, B.~McGrew, J.~Pachocki,
  A.~Petron, M.~Plappert, G.~Powell, A.~Ray, et~al.
\newblock Learning dexterous in-hand manipulation.
\newblock \emph{arXiv preprint arXiv:1808.00177}, 2018.

\bibitem[Zhu et~al.(2018)Zhu, Wang, Merel, Rusu, Erez, Cabi, Tunyasuvunakool,
  Kram{\'a}r, Hadsell, de~Freitas, et~al.]{zhu2018reinforcement}
Y.~Zhu, Z.~Wang, J.~Merel, A.~Rusu, T.~Erez, S.~Cabi, S.~Tunyasuvunakool,
  J.~Kram{\'a}r, R.~Hadsell, N.~de~Freitas, et~al.
\newblock Reinforcement and imitation learning for diverse visuomotor skills.
\newblock \emph{arXiv preprint arXiv:1802.09564}, 2018.

\bibitem[Sadeghi and Levine(2016)]{sadeghi2016cad2rl}
F.~Sadeghi and S.~Levine.
\newblock Cad2rl: Real single-image flight without a single real image.
\newblock \emph{arXiv preprint arXiv:1611.04201}, 2016.

\bibitem[Pinto et~al.(2018)Pinto, Andrychowicz, Welinder, Zaremba, and
  Abbeel]{pinto2018asym}
L.~Pinto, M.~Andrychowicz, P.~Welinder, W.~Zaremba, and P.~Abbeel.
\newblock Asymmetric actor critic for image-based robot learning.
\newblock 06 2018.

\bibitem[Tobin et~al.(2017)Tobin, Fong, Ray, Schneider, Zaremba, and
  Abbeel]{tobin2017domain}
J.~Tobin, R.~Fong, A.~Ray, J.~Schneider, W.~Zaremba, and P.~Abbeel.
\newblock Domain randomization for transferring deep neural networks from
  simulation to the real world.
\newblock In \emph{IROS}, pages 23--30. IEEE, 2017.

\bibitem[Tan et~al.(2018)Tan, Zhang, Coumans, Iscen, Bai, Hafner, Bohez, and
  Vanhoucke]{tan2018sim}
J.~Tan, T.~Zhang, E.~Coumans, A.~Iscen, Y.~Bai, D.~Hafner, S.~Bohez, and
  V.~Vanhoucke.
\newblock Sim-to-real: Learning agile locomotion for quadruped robots.
\newblock \emph{RSS}, 2018.

\bibitem[Chebotar et~al.(2018)Chebotar, Handa, Makoviychuk, Macklin, Issac,
  Ratliff, and Fox]{chebotar2018closing}
Y.~Chebotar, A.~Handa, V.~Makoviychuk, M.~Macklin, J.~Issac, N.~Ratliff, and
  D.~Fox.
\newblock Closing the sim-to-real loop: Adapting simulation randomization with
  real world experience.
\newblock \emph{arXiv preprint arXiv:1810.05687}, 2018.

\bibitem[Warde-farley et~al.(2018)Warde-farley, Wiele, Kulkarni, Ionescu,
  Hansen, and Mnih]{Wardefarley2018}
D.~Warde-farley, T.~V.~D. Wiele, T.~Kulkarni, C.~Ionescu, S.~Hansen, and
  V.~Mnih.
\newblock {Unsupervised Control through Non-parameteric Discriminative
  Rewards}.
\newblock \emph{ICLR}, pages 1--17, 2018.

\bibitem[Nair et~al.(2018)Nair, Pong, Dalal, Bahl, Lin, and
  Levine]{nair2018visual}
A.~V. Nair, V.~Pong, M.~Dalal, S.~Bahl, S.~Lin, and S.~Levine.
\newblock Visual reinforcement learning with imagined goals.
\newblock In \emph{NeurIPS}, pages 9191--9200, 2018.

\bibitem[Finn et~al.(2016)Finn, Tan, Duan, Darrell, Levine, and
  Abbeel]{finn2016deep}
C.~Finn, X.~Y. Tan, Y.~Duan, T.~Darrell, S.~Levine, and P.~Abbeel.
\newblock Deep spatial autoencoders for visuomotor learning.
\newblock In \emph{ICRA}, pages 512--519. IEEE, 2016.

\bibitem[Yu et~al.(2019)Yu, Shevchuk, Sadigh, and Finn]{yu2019unsupervised}
T.~Yu, G.~Shevchuk, D.~Sadigh, and C.~Finn.
\newblock Unsupervised visuomotor control through distributional planning
  networks.
\newblock \emph{Robotics: Science and Systems}, 2019.

\bibitem[Agrawal et~al.(2016)Agrawal, Nair, Abbeel, Malik, and
  Levine]{agrawal2016learning}
P.~Agrawal, A.~V. Nair, P.~Abbeel, J.~Malik, and S.~Levine.
\newblock Learning to poke by poking: Experiential learning of intuitive
  physics.
\newblock In \emph{NIPS}, pages 5074--5082, 2016.

\bibitem[Finn et~al.(2016)Finn, Goodfellow, and Levine]{finn2016unsupervised}
C.~Finn, I.~Goodfellow, and S.~Levine.
\newblock Unsupervised learning for physical interaction through video
  prediction.
\newblock In \emph{Advances in neural information processing systems}, pages
  64--72, 2016.

\bibitem[Finn and Levine(2017)]{finn2017deep}
C.~Finn and S.~Levine.
\newblock Deep visual foresight for planning robot motion.
\newblock In \emph{Robotics and Automation (ICRA), 2017 IEEE International
  Conference on}, pages 2786--2793. IEEE, 2017.

\bibitem[Ebert et~al.(2017)Ebert, Finn, Lee, and Levine]{ebert2017self}
F.~Ebert, C.~Finn, A.~X. Lee, and S.~Levine.
\newblock Self-supervised visual planning with temporal skip connections.
\newblock \emph{arXiv preprint arXiv:1710.05268}, 2017.

\bibitem[Ebert et~al.(2018{\natexlab{a}})Ebert, Dasari, Lee, Levine, and
  Finn]{ebert2018robustness}
F.~Ebert, S.~Dasari, A.~X. Lee, S.~Levine, and C.~Finn.
\newblock Robustness via retrying: Closed-loop robotic manipulation with
  self-supervised learning.
\newblock \emph{arXiv preprint arXiv:1810.03043}, 2018{\natexlab{a}}.

\bibitem[Ebert et~al.(2018{\natexlab{b}})Ebert, Finn, Dasari, Xie, Lee, and
  Levine]{ebert2018visual}
F.~Ebert, C.~Finn, S.~Dasari, A.~Xie, A.~Lee, and S.~Levine.
\newblock Visual foresight: Model-based deep reinforcement learning for
  vision-based robotic control.
\newblock \emph{arXiv preprint arXiv:1812.00568}, 2018{\natexlab{b}}.

\bibitem[Sermanet et~al.(2018)Sermanet, Lynch, Chebotar, Hsu, Jang, Schaal,
  Levine, and Brain]{sermanet2018time}
P.~Sermanet, C.~Lynch, Y.~Chebotar, J.~Hsu, E.~Jang, S.~Schaal, S.~Levine, and
  G.~Brain.
\newblock Time-contrastive networks: Self-supervised learning from video.
\newblock In \emph{ICRA}, pages 1134--1141. IEEE, 2018.

\bibitem[Sermanet et~al.(2016)Sermanet, Xu, and
  Levine]{sermanet2016unsupervised}
P.~Sermanet, K.~Xu, and S.~Levine.
\newblock Unsupervised perceptual rewards for imitation learning.
\newblock \emph{arXiv preprint arXiv:1612.06699}, 2016.

\bibitem[Peng et~al.(2018)Peng, Kanazawa, Toyer, Abbeel, and
  Levine]{peng2018variational}
X.~B. Peng, A.~Kanazawa, S.~Toyer, P.~Abbeel, and S.~Levine.
\newblock Variational discriminator bottleneck: Improving imitation learning,
  inverse rl, and gans by constraining information flow.
\newblock \emph{arXiv preprint arXiv:1810.00821}, 2018.

\bibitem[Finn et~al.(2017)Finn, Yu, Zhang, Abbeel, and Levine]{finn2017one}
C.~Finn, T.~Yu, T.~Zhang, P.~Abbeel, and S.~Levine.
\newblock One-shot visual imitation learning via meta-learning.
\newblock \emph{arXiv preprint arXiv:1709.04905}, 2017.

\bibitem[Huang et~al.(2015)Huang, Pan, Mulcaire, and
  Abbeel]{huang2015leveraging}
S.~H. Huang, J.~Pan, G.~Mulcaire, and P.~Abbeel.
\newblock Leveraging appearance priors in non-rigid registration, with
  application to manipulation of deformable objects.
\newblock In \emph{IROS}, pages 878--885. IEEE, 2015.

\bibitem[Lee et~al.(2015)Lee, Goldstein, Barratt, and Abbeel]{lee2015non}
A.~X. Lee, M.~A. Goldstein, S.~T. Barratt, and P.~Abbeel.
\newblock A non-rigid point and normal registration algorithm with applications
  to learning from demonstrations.
\newblock In \emph{ICRA}, pages 935--942, 2015.

\bibitem[Schulman et~al.(2013)Schulman, Lee, Ho, and
  Abbeel]{schulman2013tracking}
J.~Schulman, A.~Lee, J.~Ho, and P.~Abbeel.
\newblock Tracking deformable objects with point clouds.
\newblock In \emph{ICRA}, pages 1130--1137. IEEE, 2013.

\bibitem[Wang et~al.(2011)Wang, Miller, Fritz, Darrell, and
  Abbeel]{wang2011perception}
P.~C. Wang, S.~Miller, M.~Fritz, T.~Darrell, and P.~Abbeel.
\newblock Perception for the manipulation of socks.
\newblock In \emph{IROS 2011}, pages 4877--4884. IEEE, 2011.

\bibitem[Javdani et~al.(2011)Javdani, Tandon, Tang, O'Brien, and
  Abbeel]{javdani2011modeling}
S.~Javdani, S.~Tandon, J.~Tang, J.~F. O'Brien, and P.~Abbeel.
\newblock Modeling and perception of deformable one-dimensional objects.
\newblock In \emph{ICRA}, pages 1607--1614. IEEE, 2011.

\bibitem[Miller et~al.(2011)Miller, Fritz, Darrell, and
  Abbeel]{miller2011parametrized}
S.~Miller, M.~Fritz, T.~Darrell, and P.~Abbeel.
\newblock Parametrized shape models for clothing.
\newblock In \emph{ICRA}, pages 4861--4868. IEEE, 2011.

\bibitem[Cusumano-Towner et~al.(2011)Cusumano-Towner, Singh, Miller, O'Brien,
  and Abbeel]{cusumano2011bringing}
M.~Cusumano-Towner, A.~Singh, S.~Miller, J.~F. O'Brien, and P.~Abbeel.
\newblock Bringing clothing into desired configurations with limited
  perception.
\newblock In \emph{ICRA}, pages 3893--3900. IEEE, 2011.

\bibitem[Phillips-Grafflin and Berenson(2014)]{phillips2014representation}
C.~Phillips-Grafflin and D.~Berenson.
\newblock A representation of deformable objects for motion planning with no
  physical simulation.
\newblock In \emph{ICRA 2014}, pages 98--105. IEEE, 2014.

\bibitem[Schaul et~al.(2015)Schaul, Horgan, Gregor, and
  Silver]{schaul2015universal}
T.~Schaul, D.~Horgan, K.~Gregor, and D.~Silver.
\newblock Universal value function approximators.
\newblock In \emph{International conference on machine learning}, pages
  1312--1320, 2015.

\bibitem[Florensa et~al.(2019)Florensa, Degrave, Heess, Springenberg, and
  Riedmiller]{florensa2019self}
C.~Florensa, J.~Degrave, N.~Heess, J.~T. Springenberg, and M.~Riedmiller.
\newblock Self-supervised learning of image embedding for continuous control.
\newblock 2019.

\bibitem[Lillicrap et~al.(2016)Lillicrap, Hunt, Pritzel, Heess, Erez, Tassa,
  Silver, and Wierstra]{lillicrap2015continuous}
T.~P. Lillicrap, J.~J. Hunt, A.~Pritzel, N.~Heess, T.~Erez, Y.~Tassa,
  D.~Silver, and D.~Wierstra.
\newblock Continuous control with deep reinforcement learning.
\newblock \emph{ICLR}, 2016.

\bibitem[Todorov et~al.(2012)Todorov, Erez, and Tassa]{todorov2012mujoco}
E.~Todorov, T.~Erez, and Y.~Tassa.
\newblock Mujoco: A physics engine for model-based control.
\newblock In \emph{IROS 2012}, pages 5026--5033. IEEE, 2012.

\bibitem[Andrychowicz et~al.(2018)Andrychowicz, Baker, Chociej, Jozefowicz,
  McGrew, Pachocki, Petron, Plappert, Powell, Ray, et~al.]{OpenAI2018}
M.~Andrychowicz, B.~Baker, M.~Chociej, R.~Jozefowicz, B.~McGrew, J.~Pachocki,
  A.~Petron, M.~Plappert, G.~Powell, A.~Ray, et~al.
\newblock Learning dexterous in-hand manipulation.
\newblock \emph{arXiv preprint arXiv:1808.00177}, 2018.

\end{thebibliography}


\begin{thebibliography}{1}

\bibitem{levine2016end}
Sergey Levine, Chelsea Finn, Trevor Darrell, and Pieter Abbeel.
\newblock End-to-end training of deep visuomotor policies.
\newblock {\em The Journal of Machine Learning Research}, 17(1):1334--1373,
  2016.

\bibitem{kingma2014adam}
Diederik~P Kingma and Jimmy Ba.
\newblock Adam: A method for stochastic optimization.
\newblock {\em Proceedings of the 3rd International Conference on Learning
  Representations (ICLR)}, 2014.

\bibitem{Wardefarley2018}
David Warde-farley, Tom Van~De Wiele, Tejas Kulkarni, Catalin Ionescu, Steven
  Hansen, and Volodymyr Mnih.
\newblock {Unsupervised Control through Non-parameteric Discriminative
  Rewards}.
\newblock {\em ICLR}, pages 1--17, 2018.

\bibitem{nair2018visual}
Ashvin~V Nair, Vitchyr Pong, Murtaza Dalal, Shikhar Bahl, Steven Lin, and
  Sergey Levine.
\newblock Visual reinforcement learning with imagined goals.
\newblock {\em Advances in Neural Information Processing Systems}, pages
  9191--9200, 2018.

\bibitem{yu2019unsupervised}
Tianhe Yu, Gleb Shevchuk, Dorsa Sadigh, and Chelsea Finn.
\newblock Unsupervised visuomotor control through distributional planning
  networks.
\newblock {\em Robotics: Science and Systems}, 2019.

\bibitem{yu2019dpncode}
Tianhe Yu, Gleb Shevchuk, Dorsa Sadigh, and Chelsea Finn.
\newblock Unsupervised visuomotor control through distributional planning
  networks.
\newblock \url{https://github.com/tianheyu927/dpn}, 2019.

\end{thebibliography}

\end{document}



\section{Environment Details}
During evaluation, for all environments, a binary sparse reward is given at each time step. A positive reward $R_+=1$ is given when the goal is reached, i.e. $||s_{t+1} - s_g|| \leq \epsilon$ and a negative reward $R_-=-1$ is given otherwise. Other environment details are summarized in Table \ref{tab:env_detail}.

For the RopePush environment, to save the time for computing initial and goal configuration, 10,000 initial configurations of the rope are pre-computed and cached which are later used for training.

\begin{table*}[h!]\centering
\begin{tabular}{cP{17mm}P{15mm}P{15mm}P{15mm}cc}
\toprule
Environment & Observation  Dimension & Goal  \newline Dimension & Rendered \newline Dimension & Action  \newline Dimension & Horizon (T) & $\epsilon$ (m) \\
\midrule
Reacher & 10 & 2 & 100x100x3 &2 & 50 &0.01 \\
FetchReach & 10 & 3 & 100x100x4 & 3 & 50 & 0.05 \\
FetchPush & 25 & 3 & -& 4 & 50 & 0.05 \\
RopePush & 45 & 30 & 100x100x3 & 4 & 3 & 0.1 \\
VisualReacher (Sawyer) &  - & - & 100x100x4 & 3 & 25 & 0.1 \\
\bottomrule
\end{tabular}
\caption{Summarized environment details. The observation and goal dimension are the dimensions of the low dimension state representation when available. The rendered dimensions are the dimensions of the rendered RGBD images used in the visual experiments.}
\label{tab:env_detail}
\end{table*}
\textbf{Sawyer robot experiment details}: The observation is recorded with an Intel RealSense D435 depth camera. The goal observations are sampled by moving the robot arm to a uniformly sampled location in a cuboid of diagonal length 1.3m.  The episode was considered successful if the end effector moved to within 0.1 m from goal location at the end of the episode (we used a time horizon of 25 steps). The trained policy performs position control and outputs end effector displacement within a range of -0.05m to 0.05m in each direction.

\section{Hyper-parameters}
All the experiments are run for two random seeds. The hyper-parameters of the training algorithm with indicator rewards are summarized in Table \ref{tab:hyper_params}. For all experiments with visual observation, the parameters of the convolution layers are shared among the observation input and goal input. Due to the complexity of the RopePush environment, a spatial softmax layer \cite{levine2016end} with an output size of 32 is applied before the fully connected layers. 

\begin{table*}[h!]\centering
\begin{tabular}{@{}lp{70mm}}
\toprule
Parameter & Value\\
\midrule
\textit{shared} & \\
\hspace{5mm}positive reward ($R_+$)& 1 \\
\hspace{5mm}negative reward ($R_-$) & -1 \\
\hspace{5mm}reward filtering ($q_0$) & $\frac{1}{2}\Big[R_- + \gamma R_+/(1-\gamma) + R_+/(1-\gamma)\Big]$ \\
\hspace{5mm}optimizer & Adam \cite{kingma2014adam} \\
\hspace{5mm}learning rate & 0.001 \\
\hspace{5mm}discount ($\gamma$) & $\frac{T-1}{T}$ \\
\hspace{5mm}target network smoothing ($\tau$) & 0.98 \\
\hspace{5mm}nonlinearity & tanh \\
\midrule 
\textit{state observation} & \\
\hspace{5mm}replay buffer size & $10^6$ \\
\hspace{5mm}minibatch size & 256\\
\hspace{5mm}network architecture & 3 hidden layers with 256 neurons for each \\
\midrule 
\textit{visual observation} & \\
\hspace{5mm}replay buffer size & $5 \cdot 10^{3}$ \\
\hspace{5mm}minibatch size & 128 \\
\hspace{5mm}network architecture & 4 convolution layers followed by 3 hidden layers with 256 neurons for each \\
\bottomrule
\end{tabular}
\caption{Summarized hyper-parameters.}
\label{tab:hyper_params}
\end{table*}

\section{Ablation Analysis}
We show different ablations of our methods in Supplementary Figure~\ref{fig:visual_ablation_all_experiments}. We can see that, in the RopePush environment, filtering is required for the policy to learn. On the other hand, the Reacher and FetchReach environments show that balancing is required for optimal performance. In all cases, Indicator+Balance+Filter consistently performs better than all the ablated methods.  Thus, these results show that both balancing and filtering are important for optimal performance across a range of tasks.
\begin{figure*}[h!]
    \centering
    \begin{subfigure}[t]{0.32\linewidth}
        \includegraphics[width=\textwidth]{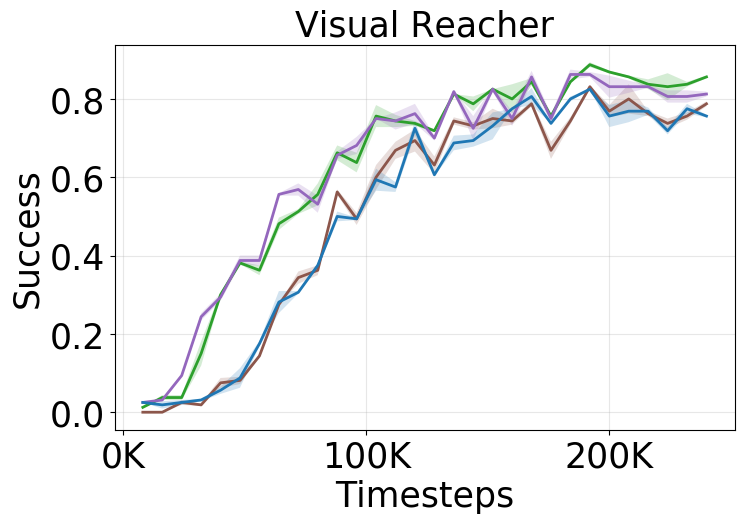}
        \label{fig:visual_Reacher_success}
    \end{subfigure}
    \begin{subfigure}[t]{0.32\linewidth}
        \includegraphics[width=\textwidth]{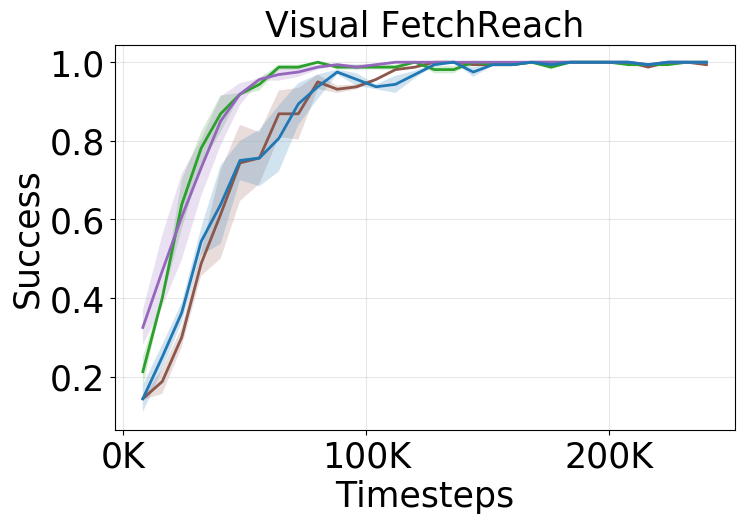}
        \label{fig:visual_FetchReach_success}
    \end{subfigure}
    \begin{subfigure}[t]{0.32\linewidth}
        \includegraphics[width=\linewidth]{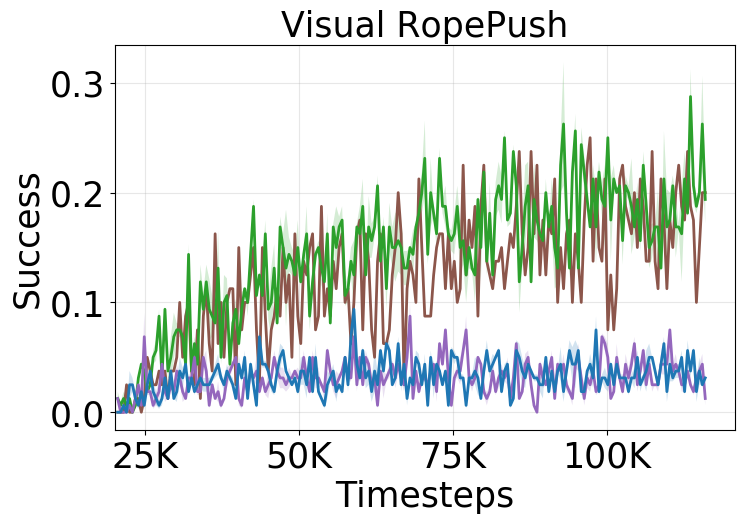}
        \label{fig:visual_RopeFloat_success}
    \end{subfigure}
    \begin{subfigure}[t]{0.32\linewidth}
        \includegraphics[width=\linewidth]{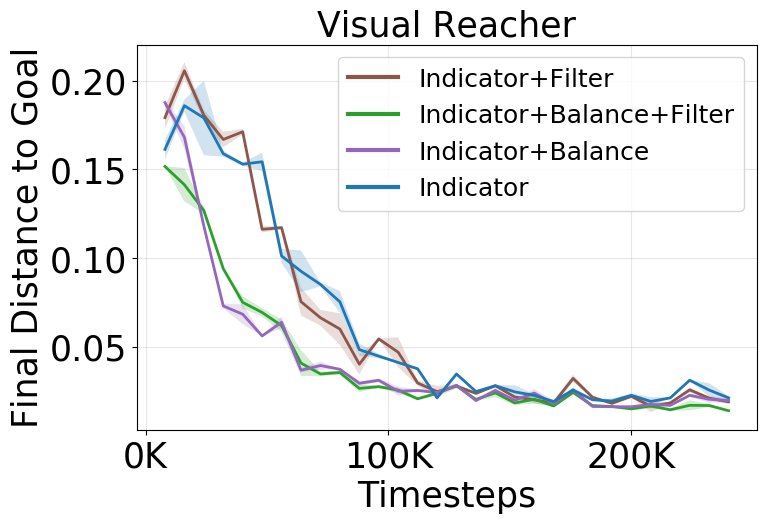}
        \label{fig:visual_Reacher_dist}
    \end{subfigure}
    \begin{subfigure}[t]{0.32\linewidth}
        \includegraphics[width=\linewidth]{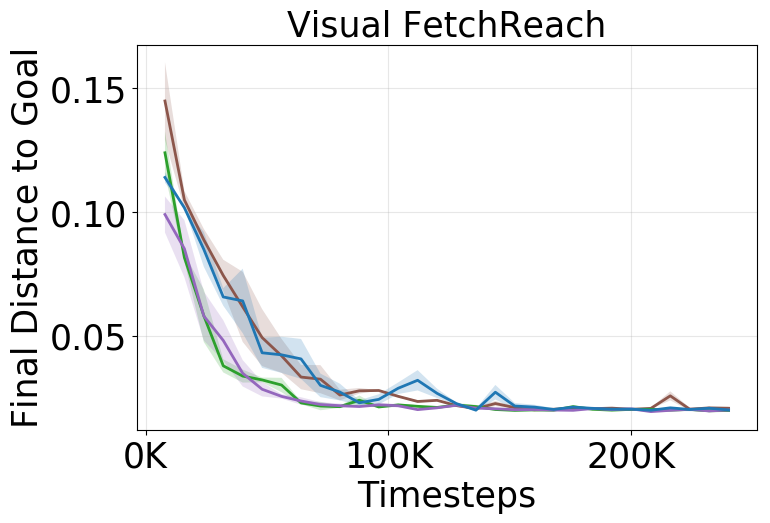}
        \label{fig:visual_FetchReach_dist}
    \end{subfigure}
    \begin{subfigure}[t]{0.32\linewidth}
        \includegraphics[width=\linewidth]{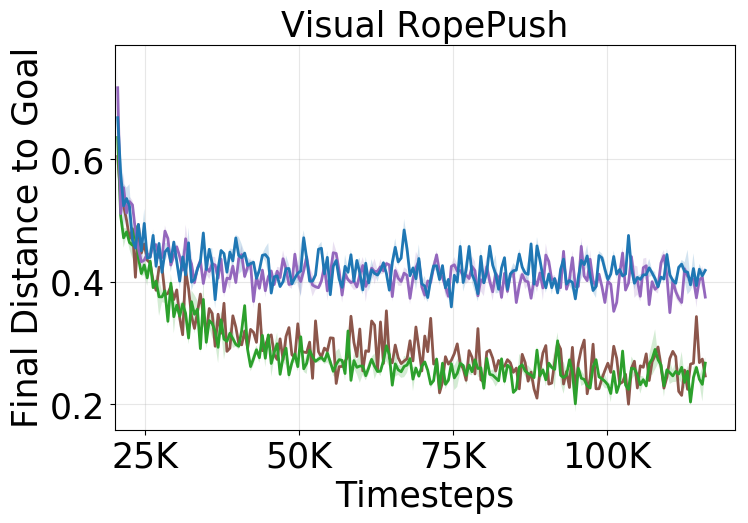}
        \label{fig:visual_RopeFloat_dist}
    \end{subfigure}
    \caption{The success (top) and the final distance to goal (bottom) of different ablations of our method.}
    \label{fig:visual_ablation_all_experiments}
\end{figure*}

With indicator rewards, we do not have any false positive rewards but may have many false negative rewards. In Supplementary Figure \ref{fig:visual_ablation_reward_accuracy}, we show the accuracy of the given rewards using different approaches with indicator rewards. The ground-truth rewards are calculated based on the ground-truth states which we do not have during training and are only used for analysis. We can see that with a longer time horizon, reward balancing is more important for the Visual Reacher and Visual FetchReach environment, which significantly lower the false negative rates. On the other hand, both reward filtering and balancing are important in the Visual RopePush environment. In all the cases, the accuracy of the rewards are improved a lot and sometimes almost perfect with reward balancing and filtering and we can see that this is necessary for the good performance of using indicator rewards.

\begin{figure*}[h!]
    \centering
    \begin{subfigure}[t]{0.32\linewidth}
        \includegraphics[width=\textwidth]{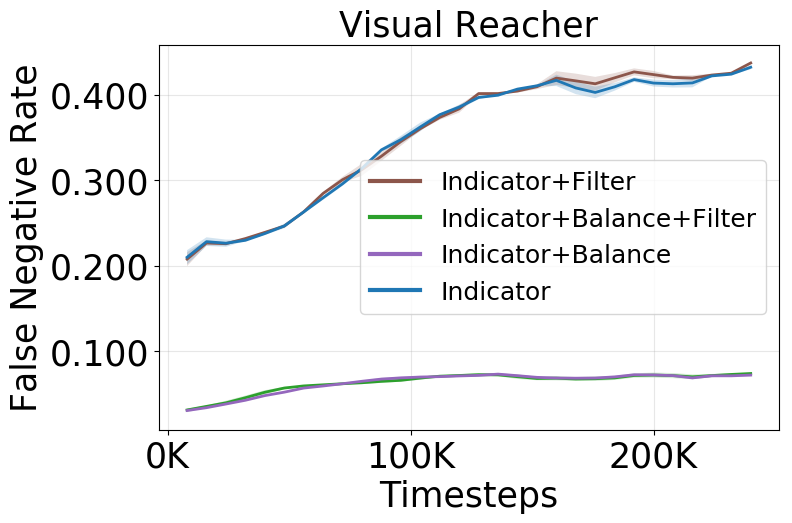}
        \label{fig:visual_Reacher_fn}
    \end{subfigure}
    \begin{subfigure}[t]{0.32\linewidth}
        \includegraphics[width=\textwidth]{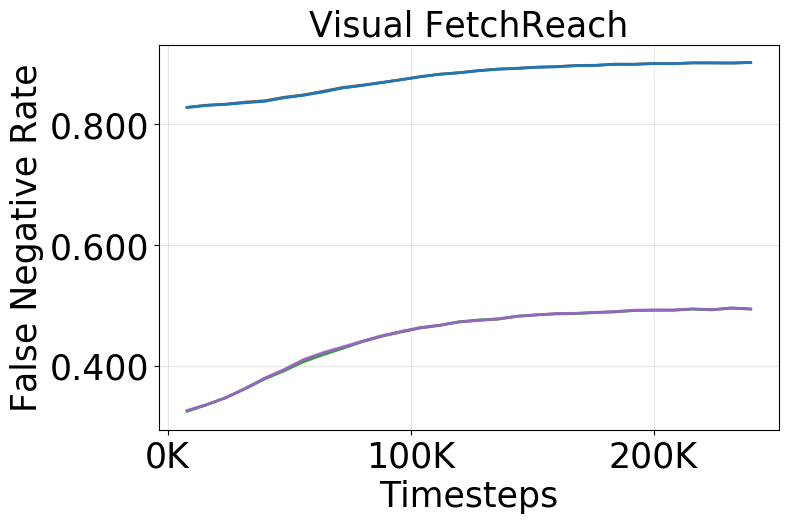}
        \label{fig:visual_FetchReach_fn}
    \end{subfigure}
    \begin{subfigure}[t]{0.32\linewidth}
        \includegraphics[width=\linewidth]{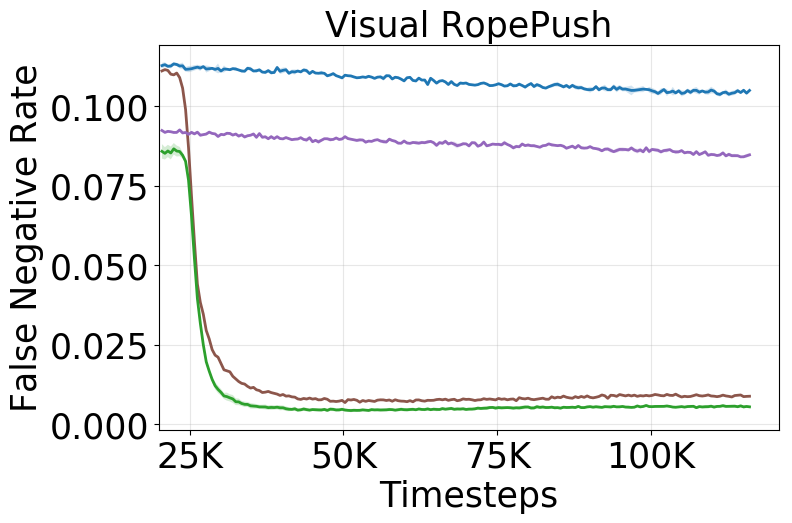}
        \label{fig:visual_RopeFloat_fn}
    \end{subfigure}
    \begin{subfigure}[t]{0.32\linewidth}
        \includegraphics[width=\linewidth]{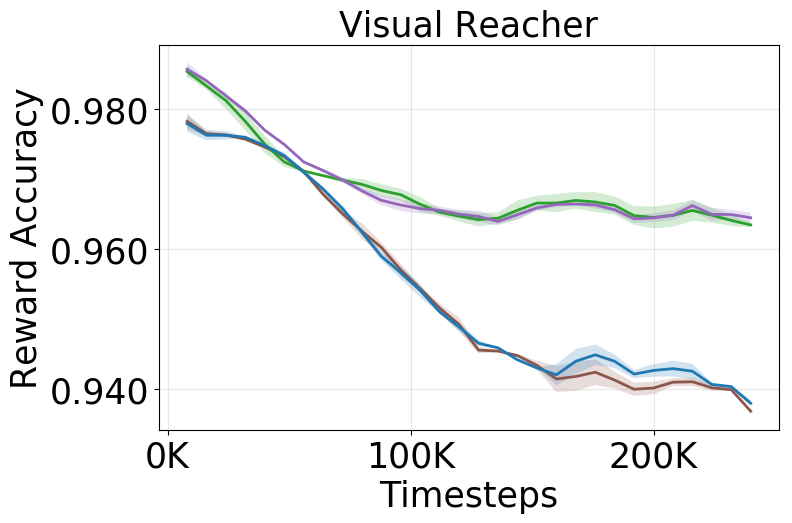}
        \label{fig:visual_Reacher_reward_accu}
    \end{subfigure}
    \begin{subfigure}[t]{0.32\linewidth}
        \includegraphics[width=\linewidth]{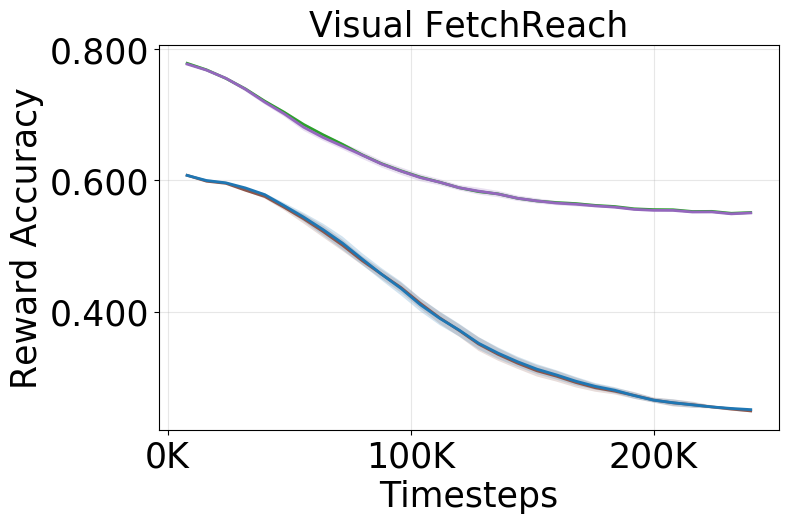}
        \label{fig:visual_FetchReach_reward_accu}
    \end{subfigure}
    \begin{subfigure}[t]{0.32\linewidth}
        \includegraphics[width=\linewidth]{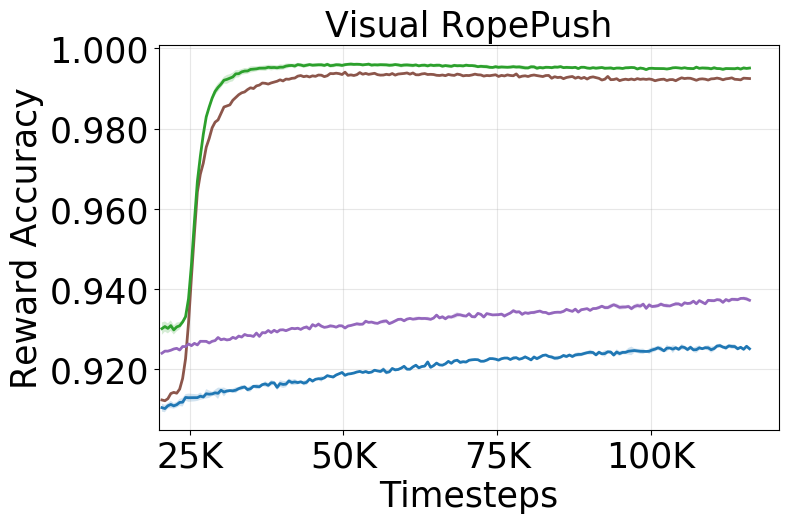}
        \label{fig:visual_RopeFloat_reward_accu}
    \end{subfigure}
    \caption{The false negative rate and the reward accuracy calculated from a batch sampled from the replay buffer of different ablations of our method.}
    \label{fig:visual_ablation_reward_accuracy}
\end{figure*}

\section{Learning with Indicator Rewards with State Input}
The performances of different learning methods when learning with low dimensional state representation are shown in Supplementary Figure \ref{fig:state_all_experiments}. In all the environments, using indicator rewards with reward balancing and filtering achieves comparable performances to Oracle. Compare this method to the Indicator baseline which does not have reward balancing and filtering, Indicator+Balance+Filter achieves a better sample efficiency. Interestingly, in the FetchReach environment, the default distance threshold for receiving an $R_+$ reward is set to 0.05. Thus, the policy that learns with this reward stops learning when the policy reach to such a distance to the goal, while the policy learned with indicator rewards keep reaching closer to the goals. This shows another benefit of using the indicator reward that the user does not need to tune the hyper-parameter $\epsilon$ to achieve the best performance. 
\begin{figure*}[h!]
    \centering
    \begin{subfigure}[t]{0.24\linewidth}
        \includegraphics[width=\textwidth]{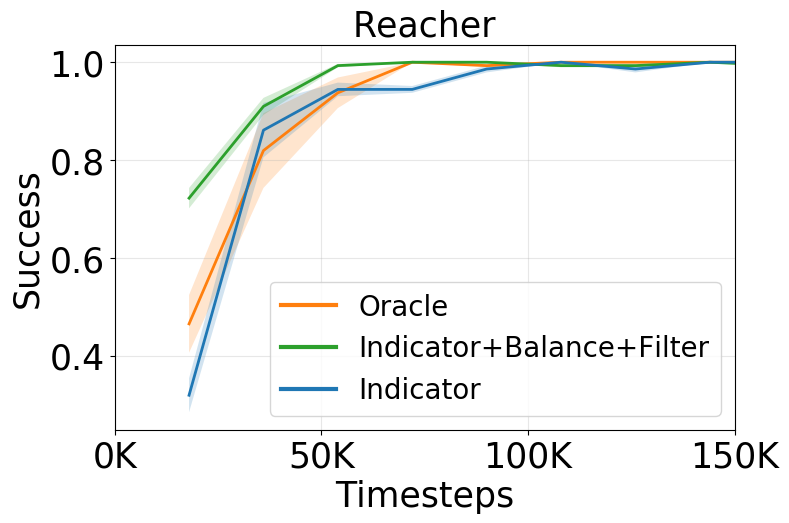}
        \label{fig:state_Reacher_success}
    \end{subfigure}
    \begin{subfigure}[t]{0.24\linewidth}
        \includegraphics[width=\textwidth]{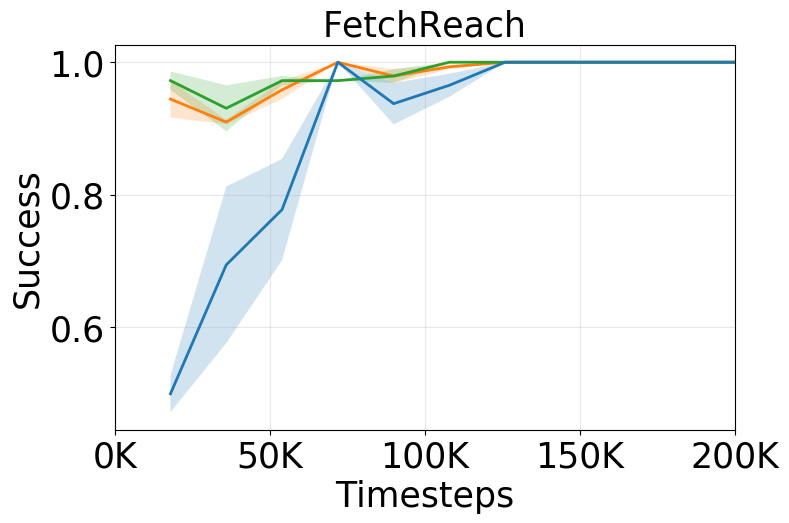}
        \label{fig:state_FetchReach_success}
    \end{subfigure}
    \begin{subfigure}[t]{0.24\linewidth}
        \includegraphics[width=\linewidth]{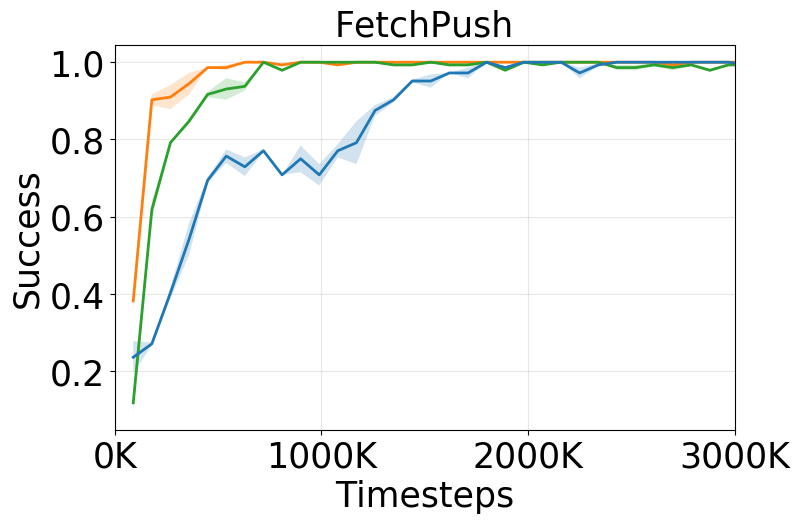}
        \label{fig:state_FetchPush_success}
    \end{subfigure}
    \begin{subfigure}[t]{0.24\linewidth}
        \includegraphics[width=\linewidth]{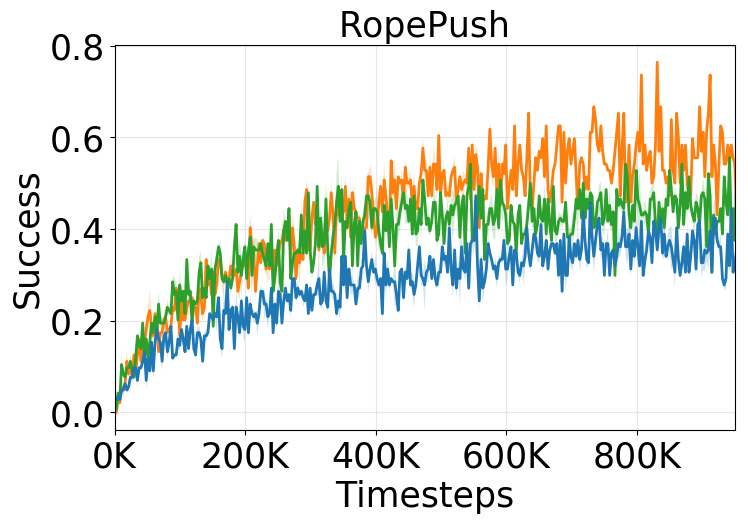}
        \label{fig:state_RopePush_success}
    \end{subfigure}
    \begin{subfigure}[t]{0.24\linewidth}
        \includegraphics[width=\textwidth]{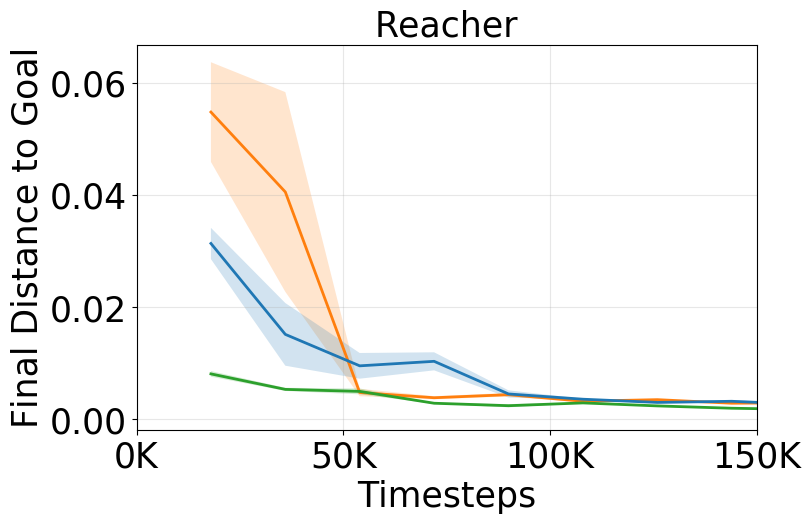}
        \label{fig:state_Reacher_dist}
    \end{subfigure}
    \begin{subfigure}[t]{0.24\linewidth}
        \includegraphics[width=\textwidth]{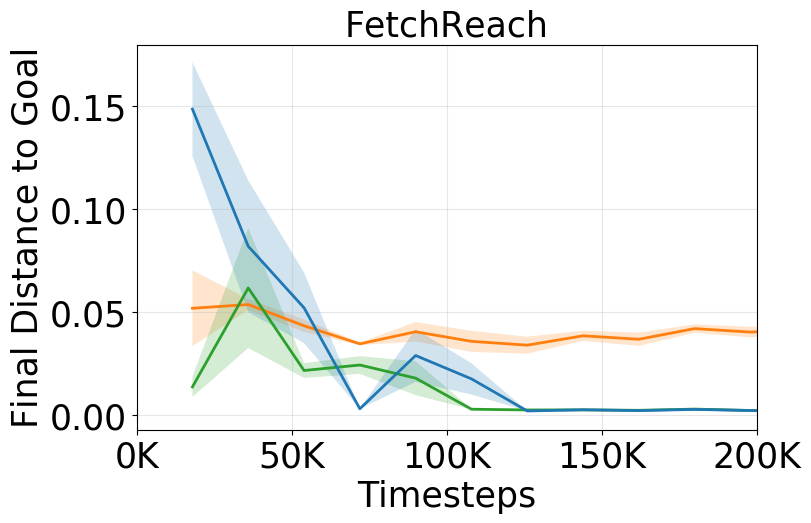}
        \label{fig:state_FetchReach_dist}
    \end{subfigure}
    \begin{subfigure}[t]{0.24\linewidth}
        \includegraphics[width=\linewidth]{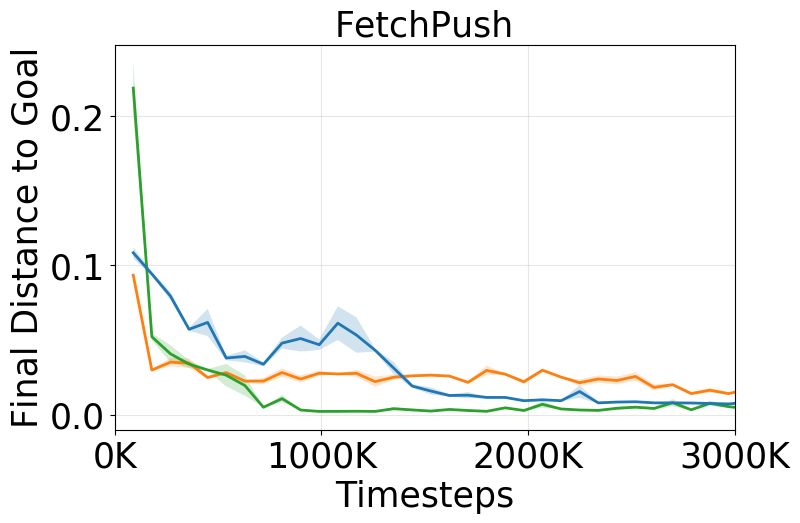}
        \label{fig:state_FetchPush_dist}
    \end{subfigure}
    \begin{subfigure}[t]{0.24\linewidth}
        \includegraphics[width=\linewidth]{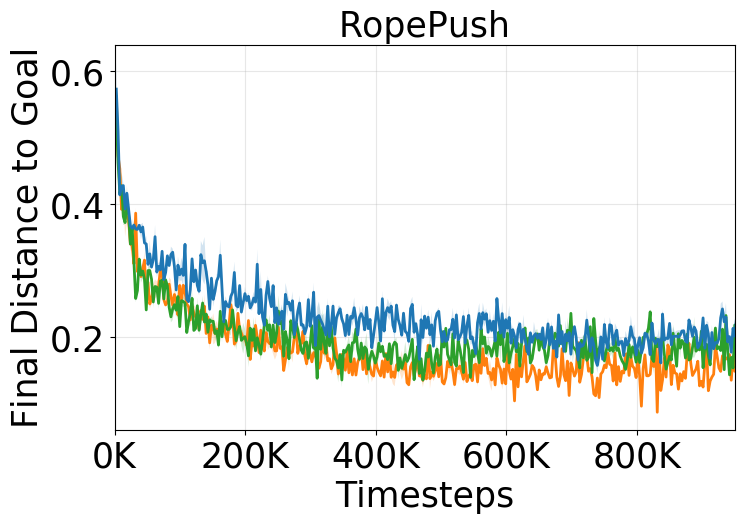}
        \label{fig:state_RopePush_dist}
    \end{subfigure}
    \caption{The success (upper row) and the final distance to goal (lower row) of different methods in different environments throughout the training. The success is defined as the mean probabily of getting the $R_+$ reward over all time steps. The final distance to goal is defined as the L2 distance to the goal in the state space, in the last time step of the episode. The input to the policy is the low dimension state representation. }
    \label{fig:state_all_experiments}
\end{figure*}

\section{Implementation Details of Baselines}
\begin{itemize}
    \item \textbf{Auto Encoder (AE)}: The network architecture for the encoder is four convolutional layers each followed by a max-pooling layer. The latent code as a dimension of 32. The decoder is four up-pooling and up-convolution layers. The auto encoder is trained jointly with the RL agent by the reconstruction loss of the observation images in the sampled transition. The learning rate is set to 0.001. We then use cosine similarity in the learned embedding space to provide dense rewards, as similarly compared in \cite{Wardefarley2018}. Specifically, assuming the learned encoding of an observation $o$ is $\phi(o)$ after $L2$ normalization, the reward will be $r(o, o_g) = max(0, \phi(o)^T\phi(o_g))$.
    \item \textbf{Variational Auto Encoder (VAE)} The VAE uses the same architecture and training procedure as the AE baseline. Following \cite{nair2018visual}, we use a variant of the VAE, $\beta$-VAE with $\beta=5$. Following \cite{nair2018visual}, given the latent representation $z, z_g$ of the current observation $o$ and the goal observation $o_g$, the rewards are given by $$ r(o, o_g) = - ||z - z_g||_2,$$ which is the negative of the Euclidean distance.
    \item \textbf{Distributional Planning Network (DPN)  \cite{yu2019unsupervised}} 
    We use the released implementation \cite{yu2019dpncode} and use the default hyperparameters. The network architecture for encoder is three convolutional layers, each with kernel size 5, stride 2. After each convolutional layer, there is layer normalization and sigmoid non-linearity. The output representation from CNN goes into a fully connected layer with latent code dimension of 128. For each environment, we collected 20,000 transitions from a random policy and performed 100,000 minibatch updates. The input dimension for each of the three environments is the same as the Rendered Dimension column specified in Table \ref{tab:env_detail}.  
    Once the DPN is pre-trained, the representation is fixed during the training of the RL agent. Following \cite{yu2019unsupervised}, given the latent representation $z, z_g$ of the current observation and the goal observation, the rewards are given by $$r(o, o_g) = -\exp(||d_\text{DPN}(z -z_g, \delta)||_1),$$ where
    $$ d_\text{DPN}(x, \delta)_i = \begin{cases} 
        \frac{1}{2}x_i^2 &  \text{for }|x_i| \leq \delta\\
        \delta |x_i| - \frac{1}{2}\delta^2 &  otherwise
    \end{cases}.$$ $\delta$ is set to 0.85. To avoid overflow of the rewards during the exponential, we normalize $z$ and $z_g$ such that they have a norm of 1 after the representation is trained.
\end{itemize}

\section{Algorithm of Learning with Indicator Rewards}
The pseudocode of our full algorithm is shown in Algorithm \ref{algo:full_algorithm}.

\begin{algorithm}[h]
\DontPrintSemicolon
$R$: Replay buffer. \;
$\pi_{\theta}$: Policy to be learned.\;
$\pi_{\beta}$: Behaviour policy.\;
$\hat{r}_{ind}: \mathcal{O} \times \mathcal{O}_g \rightarrow \mathbb{R}$: Indicator reward function \;

\For{$i \gets 1 \textrm{ to } N_{epoch}$}
{
	\For{$j \gets 1 \textrm{ to } N_{cycle}$}
	{
		Sample an initial observation $o_0$ and a goal $o_g$ \;
		Collect  $ \tau = (o_0, a_0, ..., o_T, a_T)$ following $\pi_\beta$ and goal $o_g$\;
	 	Store all transitions $(o_t, a_t, o_g, r_{t+1}, o_{t+1}, t)$ in $R$\; 
	}
	\For{$k \gets 1 \textrm{ to } N_{train}$}
	{
		Sample a mini-batch $B$ from the replay buffer\;
		\For{each transition in R}
		{
			With probability $p_1$: \qquad \tcp{$p_1 = 0.45$} 
			\Indp 
				 \label{line:st_pos_r}
					$o_g \gets o_{t+1}$ \;
					$r_{t+1} \gets \hat{r}_{ind}(o_{t+1}, o_g)$ \qquad \tcp{$r_{t+1} = R_+$} \label{line:en_pos_r}
			\Indm 
			With probability $p_2$: \qquad \tcp{$p_2 = 0.45$}
			\Indp
				Sample a future time step $t'$ from $\{t+2, ... ,T\}$\; 
				$o_g \gets o_{t'}$ \;
				$r_{t+1} \gets \hat{r}_{ind}(o_{t+1}, o_g)$ \qquad \tcp{$r_{t+1} \approx R_-$} \label{line:en_neg_r}
			\Indm
			With probability $p_3$: \qquad \tcp{$p_3 = 0.1$}
			\Indp 
				$r_{t+1} \gets \hat{r}_{ind}(o_{t+1}, o_g)$ \qquad  \tcp{$r_{t+1} \approx R_-$} \label{line:original}
			\Indm
			If $r_{t+1} = R_-$ and $Q(o_t, o_g, a_t) > q_0$:\; \label{line:filtering}
			\Indp
				Discard this transition\;
			\Indm
		}
		Perform one step of optimization using off-policy RL\;
	}
}
\caption{Learning with Indicator Reward Function}
\label{algo:full_algorithm}
\end{algorithm}

\small
\bibliographystyle{unsrt}
\bibliography{supplement}